\documentclass[letterpaper, 10 pt, conference]{ieeeconf}  

\IEEEoverridecommandlockouts                              

\usepackage{graphicx} 
\usepackage{times} 

\usepackage{times}
\usepackage{xcolor}
\usepackage{float} 
\usepackage{amsfonts}
\usepackage{amsmath}
\usepackage{multirow}
\usepackage{adjustbox}  
\usepackage{diagbox}
\usepackage[font=small]{caption}
\usepackage{tabularx}
\usepackage{url}
\usepackage{cite}

\let\labelindent\relax
\usepackage{enumitem}
\usepackage{hyperref}

\title{\LARGE \bf
\ourapproach{}: Open-Vocabulary Costmap Generation from Satellite Images and Natural Language
}

\author{
Rwik Rana\textsuperscript{1},
Jesse Quattrociocchi\textsuperscript{2},
Dongmyeong Lee\textsuperscript{1},
Christian Ellis\textsuperscript{2}, \\
Amanda Adkins\textsuperscript{1},
Adam Uccello\textsuperscript{2},
Garrett Warnell\textsuperscript{2},
Joydeep Biswas\textsuperscript{1}%
\thanks{\textsuperscript{1}The University of Texas at Austin, Austin, TX, USA.}%
\thanks{\textsuperscript{2}DEVCOM Army Research Laboratory, United States.}%
\thanks{Email: rwik2000@utexas.edu}
}

\begin{document}

\newcommand{\ourapproach}{\textsc{OVerSeeC}}

\maketitle
\thispagestyle{empty}
\pagestyle{empty}

\begin{abstract}

Aerial imagery provides essential global context for autonomous navigation, enabling route planning at scales inaccessible to onboard sensing. We address the problem of generating global costmaps for long-range planning directly from satellite imagery when entities and mission-specific traversal rules are expressed in natural language at test time. This setting is challenging since mission requirements vary, terrain entities may be unknown at deployment, and user prompts often encode compositional traversal logic. Existing approaches relying on fixed ontologies and static cost mappings cannot accommodate such flexibility. While foundation models excel at language interpretation and open-vocabulary perception, no single model can simultaneously parse nuanced mission directives, locate arbitrary entities in large-scale imagery, and synthesize them into an executable cost function for planners. We therefore propose \ourapproach{}, a zero-shot modular framework that decomposes the problem into \textit{Interpret–Locate–Synthesize}: (i) an LLM extracts entities and ranked preferences, (ii) an open-vocabulary segmentation pipeline identifies these entities from high-resolution imagery, and (iii) the LLM uses the user's natural language preferences and masks to synthesize executable costmap code. Empirically, \ourapproach{} handles novel entities, respects ranked and compositional preferences, and produces routes consistent with human-drawn trajectories across diverse regions, demonstrating robustness to distribution shifts. This shows that modular composition of foundation models enables open-vocabulary, preference-aligned costmap generation for scalable, mission-adaptive global planning. \\
Website: \texttt{\href{https://amrl.cs.utexas.edu/overseec/}{https://amrl.cs.utexas.edu/overseec/}}

\end{abstract}

\section{INTRODUCTION}

Long-range route planning for autonomous ground vehicles (AGVs) in off-road environments requires converting high-resolution aerial imagery into a planner-ready costmap. However, costmaps built on a fixed set of classes and a predefined set of traversal costs fail to adapt to mission-specific preferences, which generally have arbitrary classes.

\begin{figure}
    \centering
    \includegraphics[width=\columnwidth]{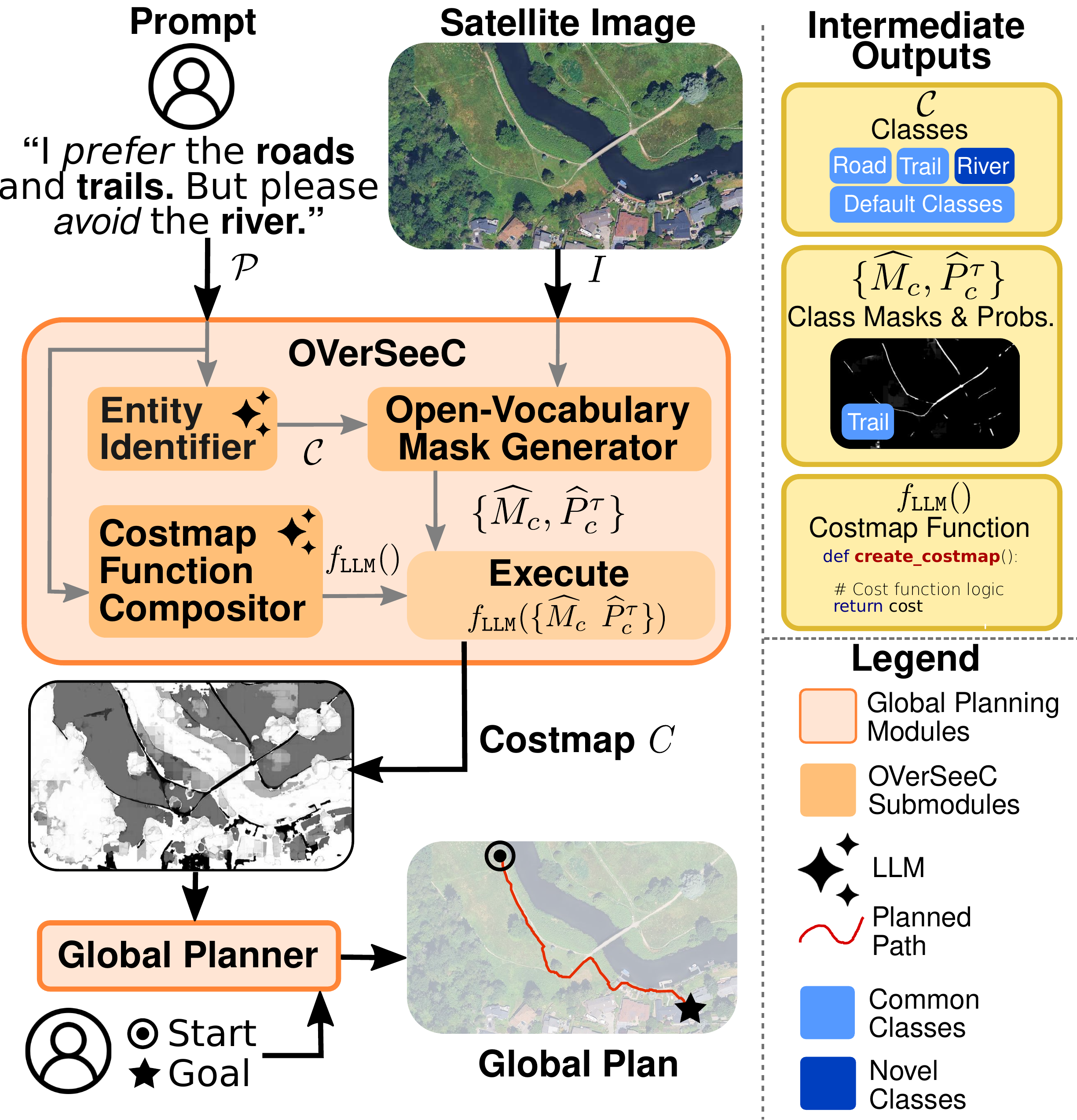}
\caption{\textbf{Overview of \ourapproach{}}, which uses a satellite image \( I \) and a natural language prompt \( \mathcal{P} \) to generate a preference-aligned costmap \( C \) for global planning. The \textit{Entity Identifier} and \textit{Costmap Function Compositor} (Sec.~\ref{sec:class_extractor},~\ref{sec:llm_code}) use an LLM to extract relevant terrain classes \( \mathcal{C} \) and synthesize a cost function \( f_{\text{LLM}}(\cdot) \) respectively. The \textit{Open-Vocabulary Mask Generator} (Sec.~\ref{sec:semseg_module},~\ref{sec:mask_refine}) performs zero-shot semantic segmentation over \( I \), yielding class masks \( \{ \widehat{M}_c \} \) and thresholded probability maps \( \{ \widehat{P}^{\tau}_c \} \), where \( c \in \mathcal{C} \). Finally \( f_{\text{LLM}}(\cdot) \) is executed to generate the final costmap \( C \).}
    \label{fig:overview}
    \vspace{-1.8em}
\end{figure}

Robust solutions exist for on-road settings, which benefit from fixed map ontologies - predefined sets of classes like road, lane marking, and sidewalk. Whereas off‑road settings present two major challenges: (i) adapting to new ontological elements, such as unencountered terrain types; (ii) adhering to complex user traversal rules and preferences (e.g., “prefer grass unless it borders a building”) \cite{viswanath2025trailblazerlearningoffroadcostmaps, mao2025pacer, skiland2022vrlpap, machines12010031}. Traditional perception models are limited by fixed ontologies in training data and thus cannot recognize novel entities at test time. Furthermore, traditional costmap generation pipelines, which rely on fixed class-to-cost mappings, do not capture the compositional and spatial logic from user preferences.

To address these challenges, we introduce \ourapproach{}, a modular, open-vocabulary, zero-shot costmap generation pipeline using aerial imagery and guided by natural language. This enables any standard global planner to compute a final route using the generated costmap. 
A single, end-to-end model is not suited to perform the distinct operations of interpreting compositional language and segmenting arbitrary entities from pixel data. Furthermore, the high resolution of satellite imagery prevents its direct processing by vision foundation models due to their fixed input-size constraints, demanding a specialized mechanism that can operate on the image at its native scale. \ourapproach{} leverages a modular design to enable decomposition of the costmap generation problem into a logical sequence—\textit{Interpret, Locate, and Synthesize}. Each stage is delegated to a specialized component: (i) a Large Language Model (LLM) for semantic entity interpretation; (ii) a perception pipeline to locate entities within high-resolution imagery; and (iii) an LLM-driven code compositor that generates a costmap function which maps the entities and the user’s compositional preferences to a costmap tailored to the mission.

Our work makes the following contributions:

\begin{enumerate}

    \item We design a zero-shot perception pipeline for high-resolution satellite imagery that performs open-vocabulary segmentation while preserving native map resolution. This enables \textit{locating} arbitrary, novel terrain classes at scale despite the fixed input-size constraints of segmentation models.
    
    \item We demonstrate that a Large Language Model (LLM) can \textit{interpret} entities and traversal rules from the user's prompt, and \textit{synthesize} executable costmap functions aligned to these preferences. This enables open-vocabulary, preference-aligned costmap generation directly from natural language.
    
    \item We develop a GUI which enables rapid, zero-shot iteration: operators can modify entities or traversal preferences in natural language and obtain updated costmaps within minutes, without annotation, retraining, or dataset-specific supervision.
    
    \item We propose the Ranked Regret Path Integral (RRPI), a metric for quantifying how well planned paths align with user preferences, enabling systematic evaluation of preference alignment.

\end{enumerate}
\section{Related Work}

Generating costmaps from satellite imagery that adapt to novel terrain and natural language preferences requires integrating solutions from several domains: remote semantic scene understanding, preference interpretation, and flexible map representations. Prior work falls into three broad strategies. (i) Semantics-first fixed-ontology methods ~\cite{ronneberger2015unet,chen2018deeplab,he2017mask, xie2021segformer, matterport_maskrcnn_2017} assume a fixed ontology of terrain classes (e.g., road, water, building) and directly train a segmentation model to predict those classes, limiting adaptability to unseen entities or dynamic user preferences. (ii) Representation learning approaches ~\cite{mao2025pacer,skiland2022vrlpap, viswanath2025trailblazerlearningoffroadcostmaps} can learn complex functions but typically require extensive training data and offer limited interpretability. (iii) Modular open-vocabulary systems adapt to novel entities and instructions without retraining, and their staged design makes each step interpretable and debuggable, making them well-suited where labeled data is scarce. The modularity facilitates future component-wise upgrades. \ourapproach{} aligns with this third strategy, using open-vocabulary VLMs\cite{luddecke2022clipseg, zhou2022maskclip, li2022languagedriven} for text-prompted segmentation, prompt-based segmentation foundation models \cite{kirillov2023segany, pointrend, densecrf, Lin2025SAMRefinerTS} for precise mask generation, and LLMs for code synthesis from natural language~\cite{liang2023code, driess2023palmeembodiedmultimodallanguage}.

Approaches such as Text2Seg~\cite{zhang2024text2segremotesensingimage} leverage CLIP-based \cite{CLIP} text embeddings for remote sensing segmentation with limited supervision. Our work extends these ideas by integrating open-vocabulary segmentation into a broader training-free costmap generation pipeline that also incorporates LLM-based preference composition, translating natural language instructions directly into executable costmap logic, thereby creating an adaptable and interpretable framework.

\section{Problem Formulation}
\label{sec:problem_formulation}
Our objective is to synthesize a scalar-valued costmap $C$ from a satellite image $I$ based on a user's natural language prompt $\mathcal{P}$, via a mapping function $f$, without requiring any task-specific training or extensive manual rule design. Formally, given $I$ and $\mathcal{P}$,
\begin{equation}
C = f\left(I, \mathcal{P}\right)
\label{eq:costmap_gen}
\end{equation}

In this formulation, $I \in \mathbb{R}^{H \times W \times 3}$ represents the input high-resolution RGB satellite image of dimensions $H \times W$. $\mathcal{P}$ is the user's natural language prompt, which can describe complex ontological and compositional preferences such as ``go over the trail, but avoid the puddle''. The function $f$ represents a mapping from $I$ and $\mathcal{P}$ as inputs to a scalar costmap $C \in [0, 1]^{H \times W}$, spatially aligned with $I$, where lower values indicate more desirable regions for traversal. 

\begin{figure*}           
  \centering
  \includegraphics[width=0.95\textwidth]{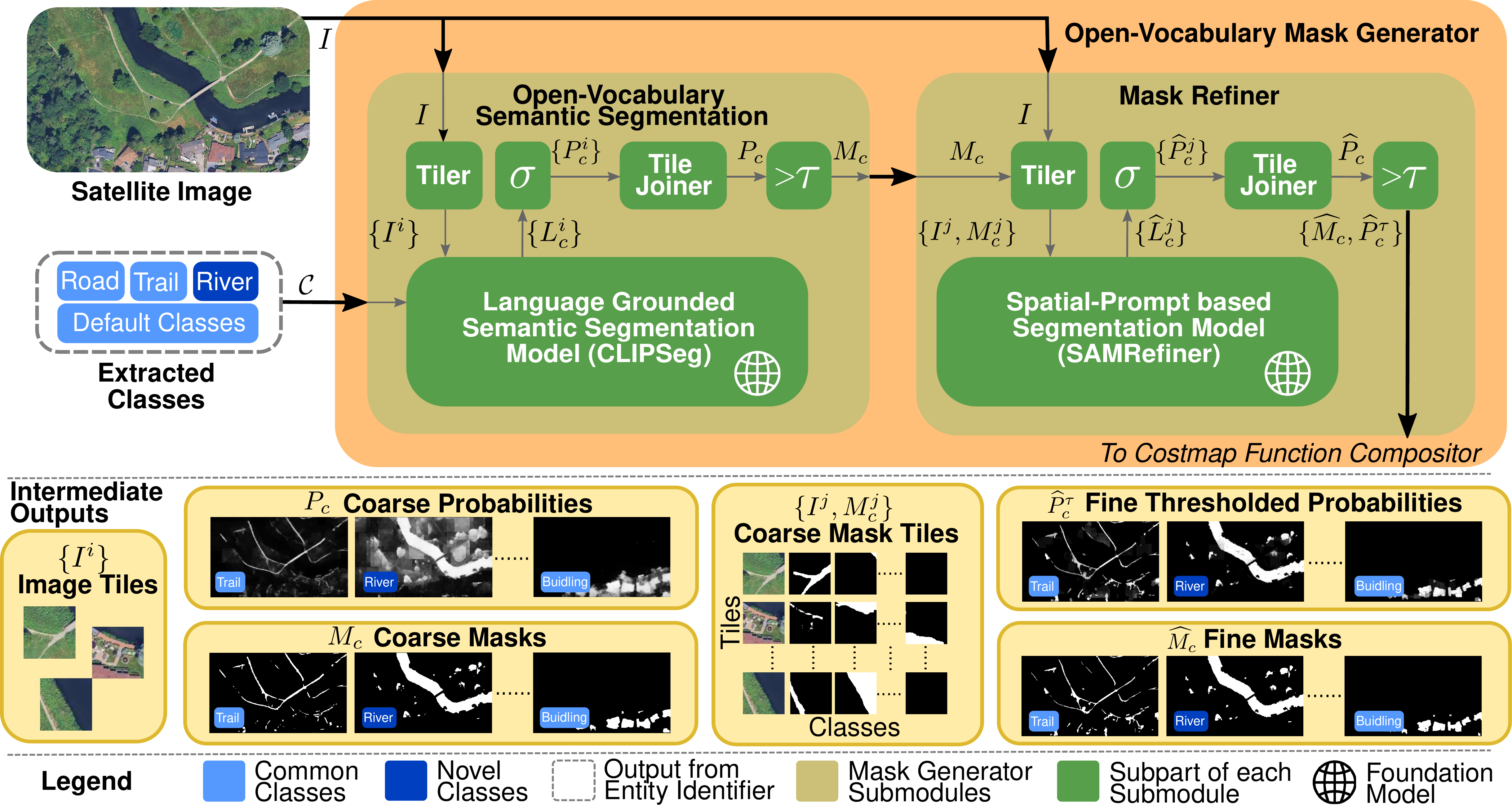}
\caption{
\textbf{Open-Vocabulary Mask Generator} (Sec.~\ref{sec:ovmg}). Given a satellite image \(I\) and extracted classes \(\mathcal{C}\), the pipeline comprises two submodules: (i) \emph{Open-Vocabulary Semantic Segmentation} (Sec.~\ref{sec:semseg_module}), which produces per-class probability maps \(P_c\) and coarse masks \(M_c\) for open-ontology classes; and (ii) \emph{Mask Refinement} (Sec.~\ref{sec:mask_refine}), which refines them into fine probabilities \(\widehat{P}_c\) and masks \(\widehat{M}_c\).}

  \label{fig:ovmg}
\end{figure*}

\section{The \ourapproach{} Algorithm}
We introduce \ourapproach{}, a modular framework for open-vocabulary costmap generation. \ourapproach{} is the realization of function $f$ (Eq.~\ref{eq:costmap_gen}) that maps a satellite image $I$ and a natural language prompt $\mathcal{P}$ to a final costmap $C$. This task demands: (i) \textit{heterogeneous skills}—semantic parsing, open-vocabulary visual grounding at native resolution, and symbolic cost composition require different inductive biases; (ii) \textit{zero-shot constraints}—the system would require task-specific supervision to handle unseen entities and compositional rules; and (iii) a \textit{scale-aware perception} strategy to process high-resolution imagery.

While a number of frontier models touch on parts of this problem, neither a single model nor a trivial combination of them is sufficient. For instance, presenting a state-of-the-art VLM like ChatGPT-4o with the image and prompt fails because it fails to produce a quantitative, pixel-aligned cost grid required by a planner. Thus, we decompose $f$ into a three-stage sequence \textit{Interpret--Locate--Synthesize} (Fig.~\ref{fig:overview}) :

\begin{enumerate}[label=(\alph*), nosep,leftmargin=*]

    \item \textbf{Entity Identification } (Sec.~\ref{sec:class_extractor}): natural language prompts are free-form, so a fixed label set of terrain/object classes cannot be assumed; the resulting class set is open-ontology. Each prompt \(\mathcal{P}\) must be parsed to construct the relevant class set \(\mathcal{C}\). Thus, an LLM is used to generate $\mathcal{C}$ which is then used for segmentation of $I$.

    \item \textbf{Open-Vocabulary Mask Generation} (Sec.~\ref{sec:ovmg}): Locating these arbitrary classes \(\mathcal{C}\) in \(I\) requires a \emph{language-grounded} segmentation model that adapts to open-ontology prompts, followed by refinement to sharpen boundaries and preserve connectivity. This stage produces per-class refined binary masks $\{\widehat{M}_c\}$ and probability maps $\{\widehat{P}_c^{\tau}\}$ for each $c \in \mathcal{C}$, using tiled inference with blending to handle high-resolution $I$.

    \item \textbf{Costmap Function Composition} (Sec.~\ref{sec:llm_code}): Fixed mappings cannot capture conditional or compositional preferences. Because the cost function must adhere to \(\mathcal{P}\), it must be generated on the fly for each prompt query. Thus an LLM synthesizes a costmap function \(f_{\text{LLM}}\) that encodes the prompt as executable logic over \(\{\widehat{M}_c\}\) and \(\{\widehat{P}_c^{\tau}\}\), with spatial predicates. 
    
    \item Finally, we execute $f_{\text{LLM}}$ using \(\{\widehat{M}_c\}\) and \(\{\widehat{P}_c^{\tau}\}\) to yield a scalar costmap $C \in [0, 1]^{H \times W}$.
\end{enumerate}

\subsection{Entity Identification}
\label{sec:class_extractor}

This first stage must parse the user's natural language prompt ($\mathcal{P}$) to identify all relevant semantic classes. Simpler methods like keyword extraction fail to handle the semantic complexity of natural language (e.g., synonyms or implicit intent), whereas an LLM can reliably identify novel classes outside any fixed ontology.

This stage also accounts for the geometric heterogeneity of these classes used for creating masks from segmentation models. Segmentation models output continuous per-pixel logit activations which are binarized to form class masks using a binarization threshold. However, activations for thin, linear structures are lower than for broad areal regions. A single high threshold fragments linear features while low values admit noise in areal masks. Consequently, classes must be grouped by geometry and assigned group-specific thresholds.

Thus, an LLM parses $\mathcal{P}$ and identifies a list of classes $\mathcal{C}$, which are then merged with a default set $\mathcal{C}_{\text{default}}$ for robust coverage. The LLM categorizes each class as either \texttt{linear} or \texttt{areal}. This distinction allows us to apply class-specific thresholds in the mask generation step (Sec.~\ref{sec:ovmg}).

\subsection{Open-Vocabulary Mask Generation}
\label{sec:ovmg}
Generating class-specific masks from high-resolution satellite imagery $I$ presents three challenges: (i) obtaining per-class semantic segmentation from open-ontology classes; (ii) refining segmentation model outputs, which generally have soft edges, speckle, and broken connectivity; and (iii) adapting to high-resolution imagery and variable image dimensions, since segmentation models accept fixed, relatively small input sizes. For (i) and (ii) we use task-specific foundation models. For (iii), we tile the images into smaller chunks and feed them to the specific models.

\subsubsection{Open-Vocabulary Semantic Segmentation}
\label{sec:semseg_module}

The first stage of our perception pipeline generates initial, coarse masks $\{M_c\}$ for each terrain category $c \in \mathcal{C}$ using a \textbf{language-grounded segmentation model (LGSM)}. This allows the system to identify arbitrary classes specified in the prompt at test time. To process the high-resolution satellite image $I$ without losing detail through downsampling, the model operates on smaller, overlapping tiles $\{I^i\}$. For each tile, LGSM produces a per-pixel probability map $P^i_c$ for every relevant class. These individual maps are then stitched together by averaging the predictions in the overlapping regions, resulting in a single, full-resolution probability maps $\{P_c\}$ for the entire area. These maps are then binarized using the class-specific (\texttt{linear} or \texttt{areal}) thresholds to produce the coarse masks $\{M_c\}$.

\subsubsection{Mask Refinement}
\label{sec:mask_refine}

These coarse masks $\{M_c\}$ often suffer from imprecise boundaries and segmentation artifacts. The second stage of the pipeline therefore refines them. This process also operates on a tiled basis, providing a \textbf{spatial prompt-based segmentation model (SPSM)} with both the original image tiles $\{I^j\}$ and the corresponding coarse masks $\{M_c^j\}$. The coarse mask acts as a strong spatial prior, guiding the SPSM to produce a significantly sharper and more accurate probability map for each tile. These refined per-tile maps are joined to form the final probability maps $\{\widehat{P}_c\}$ and its corresponding binary masks $\{\widehat{M}_c\}$. This final mask is then used to gate the probability map, yielding the thresholded probability maps $\{\widehat{P}^{\tau}_c\}$ used in the final cost composition stage.

\subsection{Costmap Function Composition}
\label{sec:llm_code}

Conventional costmap pipelines rely on a fixed class-to-cost lookup table, an approach that fails to capture conditional preferences, geometric preferences, and adapt to unseen object classes. While a simpler approach could use an LLM to generate weights for a fixed template, such a method cannot express the compositional or spatial logic inherent in user commands (e.g., ``prefer grass unless it borders a building''). We therefore propose using an LLM to synthesize executable code on-the-fly in the form of a costmap generation function, $f_{\text{LLM}}$. This ensures generality and zero-shot compositionality. The LLM translates the user's natural language prompt $\mathcal{P}$ into an executable function using the following instructions:

\begin{enumerate}[label=\roman*.]
    \item \textbf{Function signature:} Synthesize an executable function \(f_{\text{LLM}}\!:\big(\{\widehat{M}_c\},\{\widehat{P}^{\tau}_c\}\big)\!\rightarrow\! C\in[0,1]^{H\times W}\).

    \item \textbf{Mask operators:} Predefined operators are used for basic pixel-wise mask transformations—\texttt{AND}, \texttt{OR}, \texttt{NOT}, and \texttt{REMOVE} (i.e., \(\mathrm{REMOVE}(A,B)=A\land \neg B\)).

    \item \textbf{Prompt analysis:} The prompt \(\mathcal{P}\) is analyzed to (i) assign class weights \(\{w_c\}\) (lower weight indicates higher preference); (ii) detect semantic hierarchies \(\mathcal{H}\) (e.g., “baseball field” \(\subset\) “grass”); and (iii) infer geometric cues \(\{\gamma_c\}\) that map spatial language (e.g., “\textit{near} the road”) to mask transforms.

    \item \textbf{Mask operations:} An unassigned mask $\mathrm{U} \in \{0,1\}^{H\times W}$ is formed. Hierarchies \(\mathcal{H}\) are enforced by removing subset masks from parent masks to yield $\{\widehat{M}_c^{\mathcal{H}}\}$. Geometric cues \(\{\gamma_c\}\) are applied to transform $\{\widehat{M}_c^{\mathcal{H}}\}$ to geometric-aware masks $\{\widehat{M}_c^{G}\}$.

    \item \textbf{Cost accumulation:} Per-class contributions are computed as \(w_c \cdot \widehat{M}_c^{G} \cdot \widehat{P}^{\tau}_c\). The contributions are summed pixel-wise to obtain \(\tilde{C}\). Pixels in \(\mathrm{U}\) are set to \(C_{\max}=\max \tilde{C}\), yielding \(C_{\mathrm{un\text{-}normalized}}\).

    \item \textbf{Normalization:} \(C_{\mathrm{un\text{-}normalized}}\) is normalized to \([0,1]\) to yield final costmap $C$.
\end{enumerate}

\section{Implementation Details}
\label{sec:implementation_detail}

\subsection{Entity Identification \& Costmap Function Composition}
We use \texttt{gemma-2-27b-it}~\cite{gemmateam2024gemmaopenmodelsbased} as the LLM for both the entity identification (Sec.~\ref{sec:class_extractor}) and cost synthesis (Sec.~\ref{sec:llm_code}) stages. This model was chosen for its proficiency in the two capabilities required by our pipeline: (i) structured reasoning, where it accurately parses user prompts to identify and categorize entities, and (ii) robust code-generation, where it synthesizes executable Python functions to implement the user's compositional preferences.

\subsection{Open-Vocabulary Semantic Segmentation}
For the initial coarse mask generation, we select CLIPSeg~\cite{luddecke2022clipseg} as our LGSM. Its architecture is well-suited for this role, as its direct skip connections to the CLIP backbone provide strong localization for arbitrary text prompts, making it an effective open-vocabulary ``first-pass'' segmenter. As described in Sec.~\ref{sec:semseg_module}, we employ distinct thresholds for binarization. The values for linear features ($\tau_L = 0.4$) and areal features ($\tau_A = 0.8$) were chosen empirically to provide the best trade-off between preserving feature connectivity and minimizing noise across our validation datasets.

\subsection{Mask Refinement}
For the second stage of our perception pipeline, we use SAMRefiner~\cite{Lin2025SAMRefinerTS}, a variant of the Segment Anything Model (SAM)~\cite{kirillov2023segany}, as our SPSM. This choice is deliberate: SAM is a class-agnostic, prompt-based segmenter, making it the ideal tool for refining a mask when given a strong spatial prior. In our pipeline, the coarse semantic mask from CLIPSeg provides this prior.

Given the tile mask $M_c^{(i)}$ for tile $I^{(i)}$ as input, SAMRefiner internally applies a multi-prompt strategy: (i) it heuristically selects foreground and background points, (ii) incorporates the coarse mask as a Gaussian-style prior, and (iii) uses context-aware elastic bounding boxes. These prompts are passed to SAM, which generates refined masks.

\begin{figure*}[t]
  \centering
  \includegraphics[width=\textwidth]{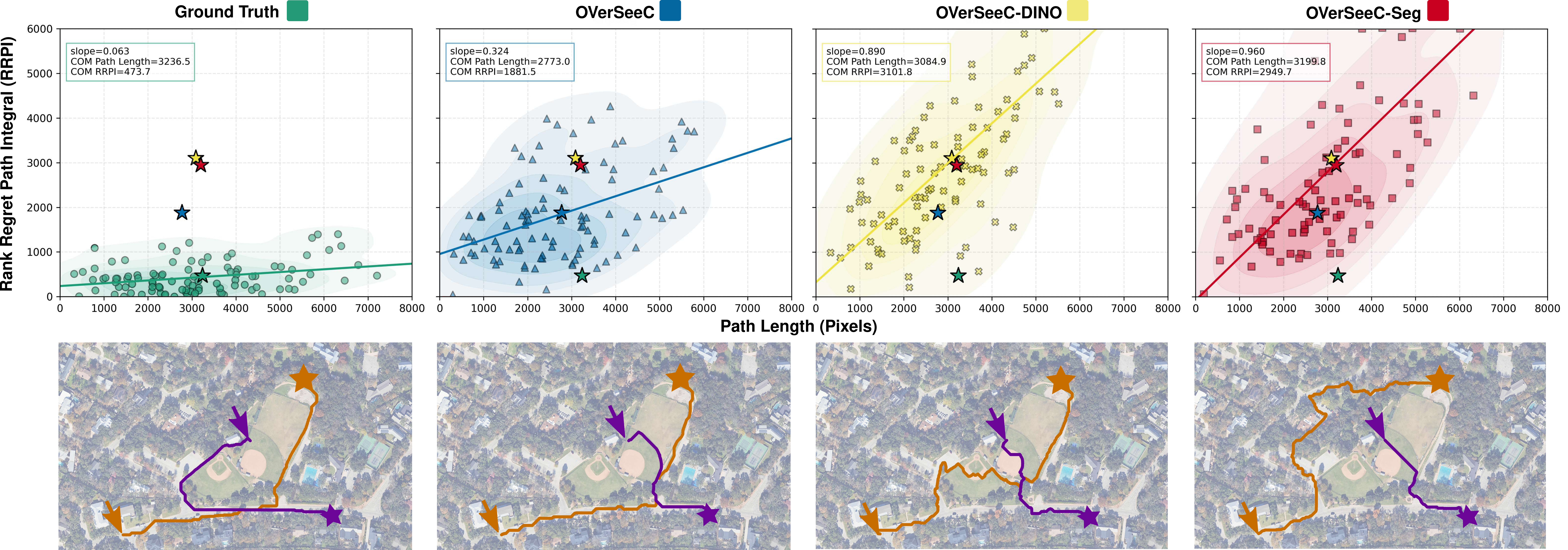}

  \caption{\textbf{Planning results for the $\mathcal{D}_2\texttt{-OOD-OV}$ scenario}:  Comparison of costmap alignment using RRPI (Sec. ~\ref{sec:rrpi_metric_definition}) metric under the user preference: \textit{“Prefer the roads and trails, grass should be fine, try to avoid the baseball field as much as possible.”} The class ranking used are: \textit{road: 1, trail: 1, grass: 2, baseball field : 3, tree: 4, building: 5}. The top row shows RRPI vs. path length scatter plots with KDE contours; the colored pointers in these plots indicates the COM of the KDE, and the solid line represents a linear regression fit. A lower slope for this line is preferable, as it indicates that the RRPI score remains low even as path length increases. The bottom row shows a subset of these trajectories generated from Dijkstra's algorithm overlaid on the map (start: arrow, goal: star).}
  \label{fig:rrpi_baseball}
  \vspace{-1.2em}
\end{figure*}

\subsection{Baselines} To benchmark against conventional semantic segmentation approaches, we use 2 fixed-ontology baselines: (i) SegFormer-B5 \cite{xie2021segformer}; (ii) DINO-UNet, which combines a frozen ViT-DINO encoder \cite{oquab2023dinov2} with a lightweight UNet decoder. We fine-tune SegFormer and train the DINO-UNet Decoder on a dataset $\mathcal{D}_1$ curated from OpenStreetMap \cite{OpenStreetMap}. It consists of image patches of size \(512\times512 \) and in total has 6000 images.

\section{Experiments and Results}
\label{sec:experiments}

Our evaluation is designed to answer three core research questions (RQs):
\begin{enumerate}
    \item \textbf{Alignment and Comparative Performance}: How well do costmaps from \ourapproach{} align with ground-truth semantic preferences, and how effectively do they guide planners to low-cost regions compared to state-of-the-art methods?
    \item \textbf{Novel-Class Generalization}: Can the zero-shot pipeline accurately segment and assign traversal costs to terrain categories mentioned in natural language prompts but absent from the supervised training ontology?
    \item \textbf{Robustness to Distribution Shift}: How well does the system maintain its segmentation accuracy and downstream planning performance with varying geographic regions or other visual domain shifts?
\end{enumerate}

\subsection{Experimental Setup}
\subsubsection{Baselines}

In all the experiments henceforth, to compare against the baselines, we will replace the LGSM i.e., CLIPSeg with baseline fixed-ontology models i.e., SegFormer and DINO-UNet, keeping all other components of \ourapproach{} as is. We name them \ourapproach{}\texttt{-Seg} and \ourapproach{}\texttt{-DINO} respectively. We use Dijkstra's algorithm to plan paths on the costmaps.

\subsubsection{Evaluation Environments}\hfill\\
We evaluate \ourapproach{} and baseline methods on two datasets, $\mathcal{D}_2$ and $\mathcal{D}_3$. 
$\mathcal{D}_2$ is designed for quantitative comparison with baselines, assessing ontological and compositional preference adherence. Ground-truth (GT) semantic maps are created by manually annotating satellite images at 0.1\,m/pixel resolution, and the dataset includes In-Distribution (ID), Out-of-Distribution (OOD), and OOD with Open-Vocabulary (OOD-OV) splits. 
$\mathcal{D}_3$ provides a more challenging setting focused on compositional preferences and human intent alignment, also at 0.1\,m/pixel resolution, and consists only of OOD and OOD-OV cases. For the human case study, maps $\mathcal{D}_3\texttt{-HE}_1$ and $\mathcal{D}_3\texttt{-HE}_2$ each use two distinct prompts, while $\mathcal{D}_3\texttt{-HE}_3$ and $\mathcal{D}_3\texttt{-HE}_4$ use three prompts each. 
Further details about both datasets are summarized in Table~\ref{tab:dataset}.

\begin{table}
\centering
\small
\renewcommand{\arraystretch}{1.15}
\label{tab:evaluation_envs}
\begin{tabular}{|l|p{6cm}|}
\hline
\textbf{Map Name} & \multicolumn{1}{|c|}{\textbf{Objective}} \\
\hline
\multicolumn{2}{|c|}{\textbf{Dataset: $\mathcal{D}_2$}} \\
\hline
\multirow{2}{*}{$\begin{aligned}
\mathcal{D}_2\texttt{-ID}_1 \\
\mathcal{D}_2\texttt{-ID}_2
\end{aligned}$}
& ID setting with familiar regions and fixed ontology from baseline training. \\
\hline
$\mathcal{D}_2\texttt{-OOD}$ & OOD region with fixed ontology. \\
\hline
$\mathcal{D}_2\texttt{-OOD-OV}$ & OOD region with prompt mentioning `baseball field' requiring OV generalization and understanding its relation to `grass'. \\
\hline
\multicolumn{2}{|c|}{\textbf{Dataset: $\mathcal{D}_3$}} \\
\hline
$\mathcal{D}_3\texttt{-HE}_1$ & Recognizing novel class `electric tower'. \\
\hline
$\mathcal{D}_3\texttt{-HE}_2$ & Differentiating `railway track' from roads. \\
\hline
$\mathcal{D}_3\texttt{-HE}_3$ & OOD region with prompt mentioning `sports fields' requiring OV generalization and understanding its relation to `grass'. \\
\hline
$\mathcal{D}_3\texttt{-HE}_4$ & Recognizing novel class `river' and its relationship to `water'. \\
\hline
\end{tabular}
\caption{Evaluation settings across environment types. \newline \textbf{ID}: In-distribution (same domain as supervised training $\mathcal{D}_1$). \textbf{OOD}: Out-of-distribution. \textbf{OV}: Open-vocabulary.}
\label{tab:dataset}
\vspace{-1.9em}
\end{table}

\subsection{Evaluation Metrics}

\subsubsection{Ranked Regret Path Integral (RRPI)}\hfill\\
\label{sec:rrpi_metric_definition}
Quantifying the alignment of a generated costmap with user preferences is challenging, as defining an ``ideal" cost function directly from natural language is often intractable. However, it is generally more straightforward to establish a preference-ordered ranking of terrain types based on a given natural language description. For instance, if a user states, ``trails are good, grass is okay, avoid water," we can assign ranks: trail (rank 1), grass (rank 2), water (rank 3), etc.

We introduce the \textbf{Ranked Regret Path Integral (RRPI)} score.
Given a user preference $\mathcal{P}$, we first derive a rank mapping $R(c)$ for each relevant semantic class $c$, where $R(c) \in \{1, 2, \dots, N_c\}$ and $N_c$ is the number of distinct classes. A lower rank indicates a more preferred terrain type, with 1 being the lowest. The \emph{rank regret} for traversing a pixel of class $c$ is $R(c) - 1$. This value penalizes less preferred terrain; the most preferred has zero regret.

For any given path $\tau = [(x_1, y_1), (x_2, y_2), \dots, (x_L, y_L)]$ of length $L$ (i.e., the trajectory covers $L$ pixels), through a semantic map $S$ (where $S(x_i, y_i)$ is the semantic class of the pixel at $(x_i, y_i)$), the RRPI score is calculated as :
\[
\text{RRPI}(\tau, S, R) = \sum_{(x,y) \in \tau} \left( R(S(x,y)) - 1 \right)
\]
where $R(S(x,y))$ is the rank of semantic class of the pixel $(x,y)$.
The RRPI score does not inherently account for path length. To provide a more nuanced evaluation, we analyze path characteristics across two dimensions: path length (in pixels) and the RRPI score ( Fig.~\ref{fig:rrpi_baseball}). We evaluate planner performance by sampling multiple start–goal pairs within the same map and preference prompt. Each planned trajectory yields a (distance, RRPI) pair, producing a set of scatter points that characterize how the planner performs under the costmap conditioned on the given image and prompt. We fit a Gaussian Kernel Density Estimate (KDE) to these scatter points. The Center of Mass (COM) of this KDE blob yields an aggregate (distance, RRPI) pair that represents the performance of the method in that specific environment for the given preference prompt. For both the distance and RRPI components of the COM, \text{lower} values are considered \textit{better}.

We calculate the RRPI scores of 50 start and goal point pairs drawn uniformly at random over the whole image for each method within each map of dataset $\mathcal{D}_2$.

\vspace{0.3em}
\subsubsection{Segmentation Accuracy Metrics}\hfill\\
\label{sec:seg_metrics_definition}
We report Intersection-over-Union (IoU) on the \emph{stitched} maps, comparing each method’s per-class masks to hand-drawn ground-truth labels from dataset $\mathcal{D}_2$.

\vspace{0.3em}
\subsubsection{Human Case Study}\hfill\\
\label{sec:human_eval}
We conducted a human case study on maps from $\mathcal{D}_3$ dataset, each with multiple prompts of varying complexity. Three annotators per prompt sketched start-to-goal trajectories that best satisfied the instructions, providing behavioral references. Alignment is quantified by an \emph{mean} Hausdorff distance between a path $\tau_{\text{sys}}$ generated by the system and the union of the three human paths $\bigcup_i \tau_{h_i}$,
\[
\mathrm{HD}(\tau_{\text{sys}})=
\frac{1}{|\tau_{\text{sys}}|}
\sum_{p\in\tau_{\text{sys}}}
\min_{q\in\bigcup_i \tau_{h_i}}\|p-q\|_2,
\]
then normalized by the map diagonal $\sqrt{H^{2}+W^{2}}$ for cross-map comparison. \textit{Lower} values indicate \textit{closer} adherence to the human-preferred paths.

\subsection{Evaluation}
Thus, we address each RQ, presenting dedicated quantitative and qualitative evidence.

\begin{figure}[!t]
\centering

\begin{minipage}{\linewidth}
\centering
\renewcommand{\arraystretch}{1.2}
\adjustbox{max width=\linewidth}{
\begin{tabular}{|l|l|c|c|}
\hline
\textbf{Map} & \textbf{Method} & \textbf{RRPI $\downarrow$} & \textbf{Path Length $\downarrow$} \\
\hline
\multirow{4}{*}{$\mathcal{D}_2\texttt{-ID}_1$}
& Ground Truth                 & 889.2   & 3682  \\ \cline{2-4}
& \ourapproach{}\texttt{-DINO} & 2613.3  & 4749  \\ \cline{2-4}
& \ourapproach{}\texttt{-Seg}  & 2424.6  & 4837  \\ \cline{2-4}
& \ourapproach{}               & \textbf{2379.9} & \textbf{4008} \\
\hline
\multirow{4}{*}{$\mathcal{D}_2\texttt{-ID}_2$}
& Ground Truth                 & 634.3   & 2452  \\ \cline{2-4}
& \ourapproach{}\texttt{-DINO} & 2214.2  & 4199  \\ \cline{2-4}
& \ourapproach{}\texttt{-Seg}  & \textbf{1923.8} & \textbf{3765} \\ \cline{2-4}
& \ourapproach{}               & 2118.2  & 3998  \\
\hline
\end{tabular}}
\captionof{table}{\textbf{RQ1 -} Comparing performances of \ourapproach{} and baselines using RRPI and total path lengths in ID environments. Best results among methods are in \textbf{bold}.}
\label{tab:rq1_rrpi_id}
\end{minipage}

\vspace{0.8em}

\begin{minipage}{\linewidth}
\centering
\renewcommand{\arraystretch}{1.2}
\adjustbox{max width=\linewidth}{
\begin{tabular}{|l|c|c|c|}
\hline
\textbf{Map} & \textbf{\ourapproach{}\texttt{-DINO}} & \textbf{\ourapproach{}\texttt{-Seg}} & \textbf{\ourapproach{}} \\
\hline
$\mathcal{D}_3\texttt{-HE}_1$ & 0.258 & 0.089 & \textbf{0.016} \\
\hline
$\mathcal{D}_3\texttt{-HE}_2$ & 0.027 & 0.073 & \textbf{0.021} \\
\hline
$\mathcal{D}_3\texttt{-HE}_3$ & 0.072 & 0.069 & \textbf{0.025} \\
\hline
$\mathcal{D}_3\texttt{-HE}_4$ & 0.090 & 0.051 & \textbf{0.026} \\
\hline
\end{tabular}}
\captionof{table}{\textbf{RQ1 —} Prompt-level alignment with human-drawn trajectories. The table reports the average of mean Hausdorff distance between human-drawn and generated trajectories for across all prompts per HE map. Lower values indicate better alignment.}

\label{tab:rq1_human_align}
\end{minipage}

\vspace{0.8em}

\begin{minipage}{\linewidth}
\centering
\small
\renewcommand{\arraystretch}{1.2}
\setlength{\tabcolsep}{2pt}
\newcolumntype{M}[1]{>{\centering\arraybackslash}m{#1}}

\begin{minipage}{0.48\linewidth}
\centering
\begin{tabular}{|M{1.5cm}|c|c|c|}
\hline
\multicolumn{2}{|c|}{$\mathcal{D}_3\texttt{-HE}_1$} & \multicolumn{2}{c|}{Human} \\
\cline{3-4}
\multicolumn{2}{|c|}{} & A & B \\
\hline
\multirow{2}{*}{\ourapproach{}} & A & \textbf{16.28} & 172.51 \\
\cline{2-4}
 & B & 256.26 & \textbf{54.92} \\
\hline
\end{tabular}
\end{minipage}\hfill
\begin{minipage}{0.48\linewidth}
\centering
\begin{tabular}{|M{1.5cm}|c|c|c|}
\hline
\multicolumn{2}{|c|}{$\mathcal{D}_3\texttt{-HE}_4$} & \multicolumn{2}{c|}{Human} \\
\cline{3-4}
\multicolumn{2}{|c|}{} & C & D \\
\hline
\multirow{2}{*}{\ourapproach{}} & C & \textbf{163.78} & 892.97 \\
\cline{2-4}
 & D & 351.13 & \textbf{151.92} \\
\hline
\end{tabular}
\end{minipage}

\captionof{table}{\textbf{RQ1 —} Prompt sensitivity between human and \ourapproach{} (Hausdorff distance, px); low diagonal values show alignment with the intended prompt. Prompts: A = `avoid electric tower', B = `can go under electric tower', C = `avoid river', D = `river is dry'.}

\label{tab:rq1_prompt_sensitivity}
\end{minipage}

\vspace{0.8em}

\includegraphics[width=\linewidth]{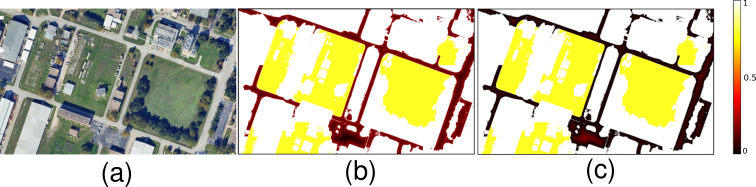}
\caption{\textbf{RQ1 —} \ourapproach{}'s alignment to geometric preferences. 
(a) is the satellite image, (b) costmap for the prompt ``stay on the center of the road,'' (c) costmap for the prompt ``stay on the side of the road,'' assigning low cost to the road edges. 
In both cases, the cost bar ranges from 0 (lowest cost) to 1 (highest).}
\label{fig:rq1_bundle}
\vspace{-1.5em}

\end{figure}

\begin{figure}[t]
\centering

\begin{minipage}{\linewidth}
\centering
\renewcommand{\arraystretch}{1.2}
\adjustbox{max width=\linewidth}{
\begin{tabular}{|l|l|c|c|}
\hline
\textbf{Map} & \textbf{Method} & \textbf{RRPI $\downarrow$} & \textbf{Path Length $\downarrow$} \\
\hline
\multirow{4}{*}{$\mathcal{D}_2\texttt{-OOD}$} 
& Ground Truth              & 249.1   & 3419  \\ \cline{2-4}
& \ourapproach{}\texttt{-DINO} & 4372.4  & 5583  \\ \cline{2-4}
& \ourapproach{}\texttt{-Seg}  & 2044.4  & \textbf{4085} \\ \cline{2-4}
& \ourapproach{}               & \textbf{1573.8} & 4351  \\
\hline
\multirow{4}{*}{$\mathcal{D}_2\texttt{-OOD-OV}$} 
& Ground Truth              & 473.7   & 3236.5 \\ \cline{2-4}
& \ourapproach{}\texttt{-DINO} & 3103.8  & 3084.9 \\ \cline{2-4}
& \ourapproach{}\texttt{-Seg}  & 2949.7  & 3199.8 \\ \cline{2-4}
& \ourapproach{}               & \textbf{1881.5} & \textbf{2773} \\
\hline
\end{tabular}}
\captionof{table}{\textbf{RQ2 and RQ3 — Novel-Class Generalization:} \ourapproach{} outperforms baselines in OOD and OOD-OV. Qualitative analysis for OOD-OV can be found in Fig.~\ref{fig:rrpi_baseball}}
\label{tab:rq2_rrpi_ov}
\end{minipage}

\vspace{0.8em}

\begin{minipage}{\linewidth}
\centering
\includegraphics[width=\linewidth]{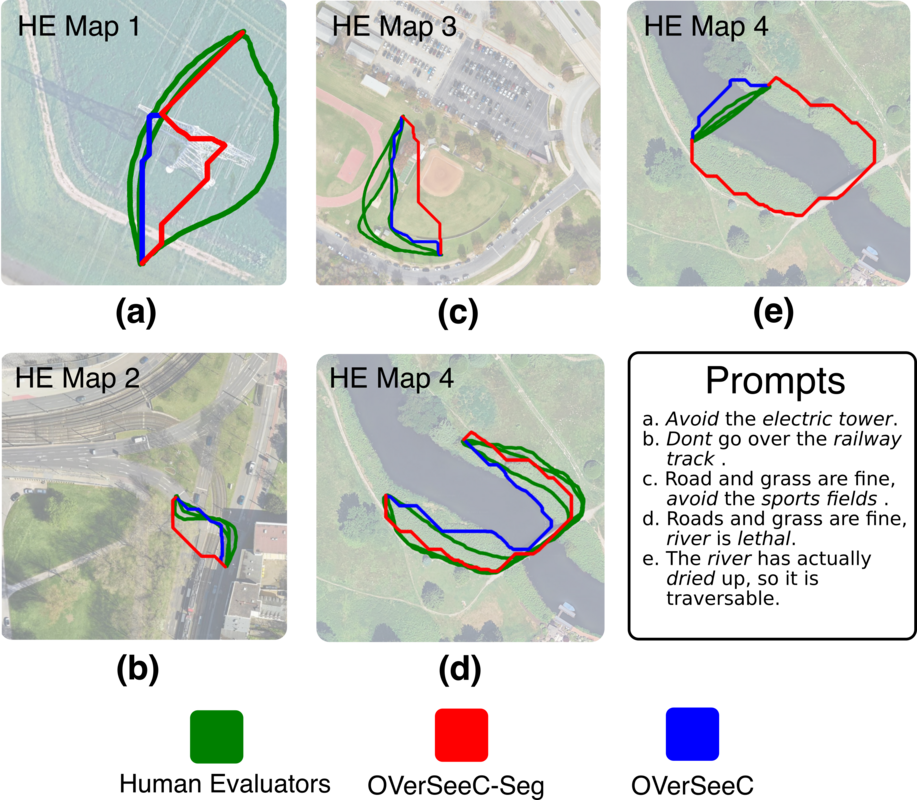}
\captionof{figure}{\textbf{RQ2 —} Qualitative results from the human case study experiments, all in open-vocabulary settings (see Table~\ref{tab:rq2_rrpi_ov}). 
Each example corresponds to the scenarios in Table~\ref{tab:rq1_human_align}, showing that \ourapproach{} adapts to novel categories and contextual prompt semantics. 
Across these OV examples, the trajectories produced by \ourapproach{} are qualitatively closest to the human-drawn references, demonstrating strong alignment with operator intent.}
\label{fig:rq2_qualitative}
\end{minipage}
\vspace{-2em}
\end{figure}

\begin{figure}[t]
\centering

\begin{minipage}{\linewidth}
\centering
\renewcommand{\arraystretch}{1.2}
\resizebox{\linewidth}{!}{%
\begin{tabular}{|l|c|c|c|c|c|}
\hline
\textbf{Method} & \textbf{Tree} & \textbf{Grass} & \textbf{Building} & \textbf{Water} & \textbf{Road/Trails} \\
\hline
\ourapproach{}\texttt{-DINO}$^{-}$ & 0.298 & 0.084 & 0.224 & 0.335 & 0.281 \\
\hline
\ourapproach{}\texttt{-DINO}       & 0.346 & 0.105 & 0.212 & 0.210 & 0.359 \\
\hline
\ourapproach{}\texttt{-Seg}$^{-}$  & 0.410 & 0.544 & 0.398 & \textbf{0.802} & 0.403 \\
\hline
\ourapproach{}\texttt{-Seg}        & 0.392 & 0.289 & 0.350 & 0.435 & 0.482 \\
\hline
\ourapproach{}$^{-}$               & \textbf{0.682} & \textbf{0.581} & 0.640 & 0.697 & 0.543 \\
\hline
\ourapproach{}                     & 0.623 & 0.517 & \textbf{0.644} & 0.665 & \textbf{0.569} \\
\hline
\end{tabular}}
\captionof{table}{\textbf{RQ3 — Segmentation Robustness:} \ourapproach{} maintains high segmentation quality (IoU) under distribution shifts. Method$^{-}$ denotes the pipeline without SAM refinement, highlighting its importance for linear features like roads.}
\label{tab:rq3_iou_ood}
\end{minipage}

\vspace{0.8em}

\begin{minipage}{\linewidth}
\centering
\includegraphics[width=\linewidth]{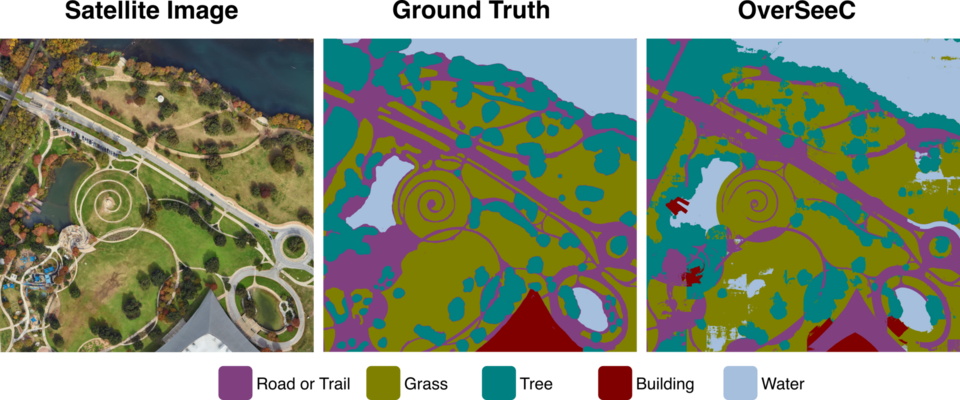}
\captionof{figure}{\textbf{RQ3 —} \ourapproach{}'s segmentation output provides a reliable foundation for planning.}
\label{fig:rq3_qualitative}
\end{minipage}
\vspace{-1.5em}
\end{figure}

\vspace{0.5em}
\subsubsection{\textbf{RQ1}: Alignment and Comparative Performance}\hfill\\
We compare \ourapproach{} to fixed-ontology baselines in $\mathcal{D}_2$ in-distribution (ID) environments. Table~\ref{tab:rq1_rrpi_id} shows that \ourapproach{} achieves competitive or better RRPI scores and path lengths, indicating strong comparative performance.

Table~\ref{tab:rq1_human_align} quantifies alignment with human-drawn trajectories across four HE maps with multiple prompts. We report the mean Hausdorff distance averaged over prompts per map, and in every case \ourapproach{} yields the lowest distance, showing closest agreement with human intent. This is reinforced by our prompt-sensitivity analysis (Table~\ref{tab:rq1_prompt_sensitivity}), which compares \ourapproach{}’s trajectories against human-drawn references under different prompts on the same map. Low diagonal values indicate close alignment with the prompt, while larger off-diagonal distances show that different instructions lead to distinct paths rather than a generic solution. 

Figure~\ref{fig:rq1_bundle} further illustrates the generated costmaps adhere to geometric preferences. For example, when the prompt specifies “stay on the center of the road” (b), the lowest cost is along the centerline, whereas with “stay on the side of the road” (c), the lowest cost shifts to the edges. In both cases, roads as a whole remain low-cost regions, confirming that \ourapproach{} can encode fine-grained geometric rules while preserving the broader semantics. 

\vspace{0.4em}
\subsubsection{\textbf{RQ2}: Novel-Class Generalization}\hfill\\
The key capability of \ourapproach{} is its ability to handle terrain classes that are unknown at deployment. We evaluate this in out-of-distribution, open-vocabulary (OOD-OV) settings, where prompts contain entities absent from the baselines’ training data. As shown in Table~\ref{tab:rq2_rrpi_ov}, \ourapproach{} achieves significantly lower RRPI than the supervised baselines. In the OOD-OV case, the baselines cannot reason about novel classes and therefore disregard the corresponding parts of the prompt, missing crucial preference information. \ourapproach{}, however, parses these entities and integrates them into the costmap, producing trajectories that reflect user intent. Figure~\ref{fig:rq2_qualitative} illustrates this capability: \ourapproach{} correctly avoids a novel electric tower, and adapts its path depending on whether a river is described as lethal or traversable, whereas baselines fail to distinguish. These results show that \ourapproach{} generalizes to open-ontology entity sets.

\vspace{0.4em}
\subsubsection{\textbf{RQ3}: Robustness to Distribution Shift}\hfill\\
Finally, we evaluate how well \ourapproach{} maintains performance when encountering geographic regions visually distinct from the training data of the supervised baselines. In these OOD settings (Table~\ref{tab:rq2_rrpi_ov}), \ourapproach{} again produces paths with lower regret, indicating its robustness to domain shift. In the OOD case, this improvement stems from the fact that fixed-ontology baselines fail to detect even known classes when appearance shifts across regions, weather, or lighting, leading to trajectories that ignore critical terrain. By contrast, \ourapproach{} leverages a language-grounded foundation model (CLIPSeg) for segmentation, trained on massive and diverse data, enabling it to consistently recognize entities, robust to such distributional shifts. This is supported by its strong underlying segmentation performance on the $\mathcal{D}_2$ dataset (Table~\ref{tab:rq3_iou_ood}), achieving higher IoU scores than the baselines on all but one class, with the largest gains observed for linear features critical to navigation.

\begin{figure}
    \centering
    \includegraphics[width=\columnwidth]{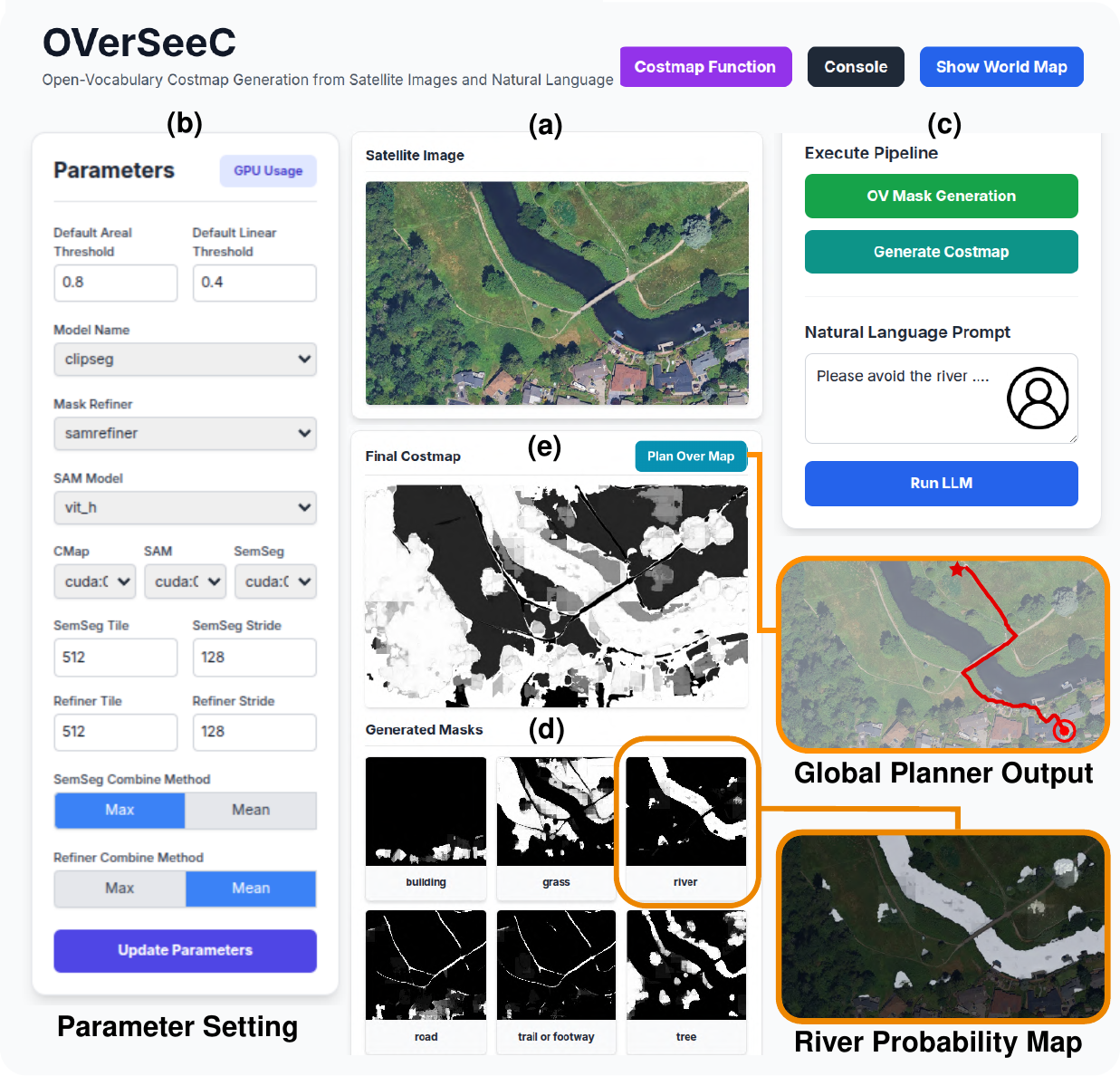}
    \caption{\textbf{\ourapproach{} GUI :} A modular interface that enables rapid, zero-shot iteration. 
    (a) Users upload a satellite image or extract imagery via the map tool. 
    (b) \ourapproach{} parameters can be adjusted. 
    (c) Natural language prompts are processed by the LLM to generate classes of interest and a costmap function. 
    (d) The open-vocabulary mask generator produces class-specific masks. 
    (e) The finalized costmap is generated and A* is used to plan over it. }
    \label{fig:gui}
    \vspace{-1.2em}
\end{figure}

\subsection{Interactive GUI for Rapid Iteration}
To demonstrate OVerSeeC’s usability as an operator-facing tool, we develop a graphical user interface (GUI) (Fig.~\ref{fig:gui}). 
The GUI showcases how operators can rapidly prototype and refine costmaps in a zero-shot manner by iterating with natural language instructions rather than retraining models. 
This design highlights OVerSeeC’s modularity and interpretability: entity masks can be generated once and reused, preferences can be updated through re-prompting or direct edits, and costmaps can be quickly validated through standard planners. 
Together, the GUI emphasizes that mission-specific costmaps can be created, inspected, and corrected within minutes, enabling practical deployment and fast operator-in-the-loop adaptation.

\section{Limitations and Future Work}

While \ourapproach{} shows strong results in zero-shot, preference-aligned costmap generation, several aspects warrant refinement. First, tighter integration between segmentation and mask refinement—e.g., through shared embeddings or joint optimization—could improve efficiency and consistency. Second, reliance on the LLM for nuanced semantic hierarchies may miss complex relationships; graph-based reasoning \cite{li2022deep} could strengthen handling of inter-related or multi-label classes. Finally, robustness to visual artifacts (e.g., shadows) and occlusions is limited, and incorporating contextual cues or generative inpainting may yield more coherent costmaps for planning.

\section{Conclusion}
We present \ourapproach{}, a modular and zero-shot architecture for generating costmaps from aerial imagery using natural language preferences, addressing the need for adaptability in off-road navigation without requiring fine-tuning.
By leveraging language-grounded segmentation, mask refinement, and LLM-driven preference interpretation, \ourapproach{} enables rapid adaptation to new classes and compositional instructions.
Empirical evaluations demonstrate its high adaptability, successful generalization to novel terrains and preferences, and superior performance over baselines in challenging out-of-distribution and open-vocabulary scenarios.
This work highlights the potential of combining large-scale pre-trained models in neuro-symbolic frameworks for creating adaptable, user-centric robotic navigation systems.

\section*{Acknowledgments}
This work was partially supported by ARL SARA (W911NF-24-2-0025, W911NF-23-2-0211). The views expressed are those of the authors and do not necessarily reflect those of the sponsors.

\bibliographystyle{unsrt}

\bibliography{references}

@InProceedings{ronneberger2015unet,
  author    = "Ronneberger, Olaf and Fischer, Philipp and Brox, Thomas",
  editor    = "Navab, Nassir and Hornegger, Joachim and Wells, William M. and Frangi, Alejandro F.",
  title     = "U-Net: Convolutional Networks for Biomedical Image Segmentation",
  booktitle = "Medical Image Computing and Computer-Assisted Intervention -- MICCAI 2015",
  year      = "2015",
  publisher = "Springer International Publishing",
  address   = "Cham",
  pages     = "234--241",
  isbn      = "978-3-319-24574-4"
}

@InProceedings{chen2018deeplab,
author = {Chen, Liang-Chieh and Zhu, Yukun and Papandreou, George and Schroff, Florian and Adam, Hartwig},
title = {Encoder-Decoder with Atrous Separable Convolution for Semantic Image Segmentation},
booktitle = {Proceedings of the European Conference on Computer Vision (ECCV)},
month = {September},
year = {2018}
}

@InProceedings{he2017mask,
author = {He, Kaiming and Gkioxari, Georgia and Dollar, Piotr and Girshick, Ross},
title = {Mask R-CNN},
booktitle = {Proceedings of the IEEE International Conference on Computer Vision (ICCV)},
month = {Oct},
year = {2017}
}

@InProceedings{luddecke2022clipseg,
    author    = {L\"uddecke, Timo and Ecker, Alexander},
    title     = {Image Segmentation Using Text and Image Prompts},
    booktitle = {Proceedings of the IEEE/CVF Conference on Computer Vision and Pattern Recognition (CVPR)},
    month     = {June},
    year      = {2022},
    pages     = {7086-7096}
}

@article{kirillov2023segany,
  title={Segment Anything},
  author={Kirillov, Alexander and Mintun, Eric and Ravi, Nikhila and Mao, Hanzi and Rolland, Chloe and Gustafson, Laura and Xiao, Tete and Whitehead, Spencer and Berg, Alexander C. and Lo, Wan-Yen and Doll{\'a}r, Piotr and Girshick, Ross},
  journal={arXiv:2304.02643},
  year={2023}
}

@inproceedings{
li2022languagedriven,
title={Language-driven Semantic Segmentation},
author={Boyi Li and Kilian Q Weinberger and Serge Belongie and Vladlen Koltun and Rene Ranftl},
booktitle={International Conference on Learning Representations},
year={2022},
url={https://openreview.net/forum?id=RriDjddCLN}
}

@INPROCEEDINGS{skiland2022vrlpap,
  author={Sikand, Kavan Singh and Rabiee, Sadegh and Uccello, Adam and Xiao, Xuesu and Warnell, Garrett and Biswas, Joydeep},
  booktitle={2022 International Conference on Robotics and Automation (ICRA)}, 
  title={Visual Representation Learning for Preference-Aware Path Planning}, 
  year={2022},
  volume={},
  number={},
  pages={11303-11309},
  doi={10.1109/ICRA46639.2022.9811828}}

@ARTICLE{mao2025pacer,
  author={Mao, Luisa and Warnell, Garrett and Stone, Peter and Biswas, Joydeep},
  journal={IEEE Robotics and Automation Letters}, 
  title={pacer: Preference-Conditioned All-Terrain Costmap Generation}, 
  year={2025},
  volume={10},
  number={5},
  pages={4572-4579},
  doi={10.1109/LRA.2025.3549645}}

@misc{gemmateam2024gemmaopenmodelsbased,
      title={Gemma: Open Models Based on Gemini Research and Technology}, 
      author={Gemma Team and Co. Authors},
      year={2024},
      eprint={2403.08295},
      archivePrefix={arXiv},
      primaryClass={cs.CL},
      url={https://arxiv.org/abs/2403.08295}, 
}

@InProceedings{zhou2022maskclip,
    author = {Zhou, Chong and Loy, Chen Change and Dai, Bo},
    title = {Extract Free Dense Labels from CLIP},
    booktitle = {European Conference on Computer Vision (ECCV)},
    year = {2022}
}

@inproceedings{densecrf,
 author = {Kr\"{a}henb\"{u}hl, Philipp and Koltun, Vladlen},
 booktitle = {Advances in Neural Information Processing Systems},
 editor = {J. Shawe-Taylor and R. Zemel and P. Bartlett and F. Pereira and K.Q. Weinberger},
 pages = {},
 publisher = {Curran Associates, Inc.},
 title = {Efficient Inference in Fully Connected CRFs with Gaussian Edge Potentials},
 url = {https://proceedings.neurips.cc/paper_files/paper/2011/file/beda24c1e1b46055dff2c39c98fd6fc1-Paper.pdf},
 volume = {24},
 year = {2011}
}

@INPROCEEDINGS{pointrend,
  author={Kirillov, Alexander and Wu, Yuxin and He, Kaiming and Girshick, Ross},
  booktitle={2020 IEEE/CVF Conference on Computer Vision and Pattern Recognition (CVPR)}, 
  title={PointRend: Image Segmentation As Rendering}, 
  year={2020},
  volume={},
  number={},
  pages={9796-9805},
  doi={10.1109/CVPR42600.2020.00982}}

@inproceedings{liang2023code,
  title={Code as policies: Language model programs for embodied control},
  author={Liang, Jacky and Huang, Wenlong and Xia, Fei and Xu, Peng and Hausman, Karol and Ichter, Brian and Florence, Pete and Zeng, Andy},
  booktitle={2023 IEEE International Conference on Robotics and Automation (ICRA)},
  pages={9493--9500},
  year={2023},
}

@misc{driess2023palmeembodiedmultimodallanguage,
      title={PaLM-E: An Embodied Multimodal Language Model}, 
      author={Danny Driess and Fei Xia and Mehdi S. M. Sajjadi and Co. Authors.},
      year={2023},
      eprint={2303.03378},
      archivePrefix={arXiv},
      primaryClass={cs.LG},
      url={https://arxiv.org/abs/2303.03378}, 
}

@inproceedings{xie2021segformer,
  title={SegFormer: Simple and Efficient Design for Semantic Segmentation with Transformers},
  author={Xie, Enze and Wang, Wenhai and Yu, Zhiding and Anandkumar, Anima and Alvarez, Jose M and Luo, Ping},
  booktitle={Neural Information Processing Systems (NeurIPS)},
  year={2021}

}

@misc{oquab2023dinov2,
  title={DINOv2: Learning Robust Visual Features without Supervision},
  author={Oquab, Maxime and Darcet, Timothée and Moutakanni and Co. Authors},
  journal={arXiv:2304.07193},
  year={2023}
}

@misc{OpenStreetMap,
   author = {{OpenStreetMap contributors}},
   title = {{Planet dump retrieved from \url{https://planet.osm.org} }},
   year = {2017},
 }

@misc{zhang2024text2segremotesensingimage,
      title={Text2Seg: Remote Sensing Image Semantic Segmentation via Text-Guided Visual Foundation Models}, 
      author={Jielu Zhang and Zhongliang Zhou and Gengchen Mai and Mengxuan Hu and Zihan Guan and Sheng Li and Lan Mu},
      year={2024},
      eprint={2304.10597},
      archivePrefix={arXiv},
      primaryClass={cs.CV},
      url={https://arxiv.org/abs/2304.10597}, 
}

@misc{matterport_maskrcnn_2017,
  title={Mask R-CNN for object detection and instance segmentation on Keras and TensorFlow},
  author={Waleed Abdulla},
  year={2017},
  publisher={Github},
  journal={GitHub repository},
  howpublished={\url{https://github.com/matterport/Mask_RCNN}},
}

@article{li2022deep,
  title={Deep Hierarchical Semantic Segmentation},
  author={Li, Liulei and Zhou, Tianfei and Wang, Wenguan and Li, Jianwu and Yang, Yi},
  journal={arXiv preprint arXiv:2203.14335},
  year={2022}
}

@article{Lin2025SAMRefinerTS,
  title={SAMRefiner: Taming Segment Anything Model for Universal Mask Refinement},
  author={Yuqi Lin and Hengjia Li and Wenqi Shao and Zheng Yang and Jun Zhao and Xiaofei He and Ping Luo and Kaipeng Zhang},
  journal={ArXiv},
  year={2025},
  volume={abs/2502.06756},
  url={https://api.semanticscholar.org/CorpusID:276249777}
}

@misc{viswanath2025trailblazerlearningoffroadcostmaps,
      title={Trailblazer: Learning offroad costmaps for long range planning}, 
      author={Kasi Viswanath and Felix Sanchez and Timothy Overbye and Jason M. Gregory and Srikanth Saripalli},
      year={2025},
      eprint={2505.09739},
      archivePrefix={arXiv},
      primaryClass={cs.RO},
      url={https://arxiv.org/abs/2505.09739}, 
}

@article{CLIP,
  author       = {Alec Radford and
                  Jong Wook Kim and
                  Chris Hallacy and
                  Aditya Ramesh and
                  Gabriel Goh and
                  Sandhini Agarwal and
                  Girish Sastry and
                  Amanda Askell and
                  Pamela Mishkin and
                  Jack Clark and
                  Gretchen Krueger and
                  Ilya Sutskever},
  title        = {Learning Transferable Visual Models From Natural Language Supervision},
  journal      = {CoRR},
  volume       = {abs/2103.00020},
  year         = {2021},
  url          = {https://arxiv.org/abs/2103.00020},
  eprinttype    = {arXiv},
  eprint       = {2103.00020},
  bibsource    = {dblp computer science bibliography, https://dblp.org}
}

@Article{machines12010031,
AUTHOR = {Wang, Nan and Li, Xiang and Zhang, Kanghua and Wang, Jixin and Xie, Dongxuan},
TITLE = {A Survey on Path Planning for Autonomous Ground Vehicles in Unstructured Environments},
JOURNAL = {Machines},
VOLUME = {12},
YEAR = {2024},
NUMBER = {1},
ARTICLE-NUMBER = {31},
URL = {https://www.mdpi.com/2075-1702/12/1/31},
ISSN = {2075-1702},
DOI = {10.3390/machines12010031}
}

\end{document}